%% file: main_arxiv.tex
\documentclass{article}

\PassOptionsToPackage{numbers,sort&compress}{natbib}

\usepackage[preprint]{corl_2026}

\usepackage[T1]{fontenc}
\usepackage{amsmath}
\usepackage{amssymb}
\usepackage{mathtools}
\usepackage{graphicx}
\usepackage{booktabs}
\usepackage{multirow}
\usepackage{url}
\usepackage{bbm}
\usepackage{xcolor}
\usepackage{enumitem}
\usepackage{siunitx}
\usepackage{wrapfig}
\usepackage{subcaption}

\usepackage{algorithm}
\usepackage{algpseudocode}

\title{From Solo to Ensemble: \\A Hierarchical Framework for Composable Multi-Agent Human-Object Interaction}

\author{
  {\normalfont Zekai Deng\textsuperscript{1}\quad
  Kangyi Chen\textsuperscript{1}\quad
  Ye Shi\textsuperscript{1,2}\quad
  Jingya Wang\textsuperscript{1,}\thanks{Corresponding author: wangjingya@shanghaitech.edu.cn}}\\[6pt]
  {\normalfont \{dengzk2023, chenky2025, shiye, wangjingya\}@shanghaitech.edu.cn}\\[6pt]
  {\normalfont \textsuperscript{1}ShanghaiTech University\quad \textsuperscript{2}InstAdapt}
}

\hypersetup{
  pdftitle={From Solo to Ensemble: A Hierarchical Framework for Composable Multi-Agent Human-Object Interaction},
  pdfauthor={Zekai Deng, Kangyi Chen, Ye Shi, Jingya Wang},
  pdfsubject={Preprint}
}

\begin{document}

\maketitle

\begin{abstract}
Physics-based human-object interaction has achieved robust single-agent manipulation skills, yet extending them to multi-agent cooperative tasks remains challenging. Existing approaches typically adapt interaction policies through task-specific fine-tuning, which entangles low-level contact-rich execution with high-level coordination and limits reuse across object geometries, interaction types, and team sizes. We propose a hierarchical framework that converts a single-agent HOI policy into a reusable \emph{Object-oriented Motion Skill}. Specifically, we reinterpret teacher rollouts as object-oriented action supervision by extracting short-horizon object-proxy motions from executed trajectories, and distill task-specific teachers into a low-level skill operating in an \emph{Object-oriented Action Space}. For downstream tasks, the distilled skill is frozen as a reusable executor, while a high-level policy coordinates multiple agents by generating region-wise object-oriented actions conditioned on the shared object, task goal, agent states, and local manipulation regions. This formulation shifts multi-agent HOI learning from direct contact-rich full-body control to compact object-level proxy-motion coordination. Experiments on diverse HOI tasks show that the distilled \emph{Object-oriented Motion Skill} supports robust proxy-motion execution and enables composable policy learning across different interaction types, object geometries, and team sizes.
\end{abstract}


\input{sec/introduction}

\input{sec/relatedwork}

\input{sec/method}

\input{sec/experiment}

\input{sec/limitation}

\input{sec/conclusion}

\acknowledgments{
This work was supported by the National Natural Science Foundation of China (62406195, 62303319), HPC Platform of ShanghaiTech University, Core Facility Platform of Computer Science and Communication of ShanghaiTech University, Key Laboratory of Intelligent Perception and Human-Machine Collaboration (ShanghaiTech University), Ministry of Education, and Shanghai Engineering Research Center of Intelligent Vision and Imaging.
}


\bibliography{references}

\clearpage
\appendix

\counterwithin{figure}{section}
\counterwithin{table}{section}
\counterwithin{equation}{section}

\input{sec/appendix}

\end{document}

%% file: sec/introduction.tex
\section{Introduction}

Physics-based character control has made substantial progress in learning robust full-body skills from motion imitation, adversarial motion priors, latent skill learning, and unified humanoid controllers~\citep{peng2018deepmimic,peng2021amp,peng2022ase,tessler2023calm,tessler2024maskedmimic}. Recent human-object interaction (HOI) methods further extend these advances to contact-rich object manipulation and human-scene interaction~\citep{wang2023physhoi,wang2024skillmimic,tevet2024closd,pan2025tokenhsi,xu2025intermimic}. However, multi-human HOI introduces a distinct coordination challenge: multiple humanoids must coordinate contact locations, motion directions, and force application around the same manipulated object.

Existing methods have explored transferring interaction skills to multi-human tasks. 
For example, CooHOI~\citep{gao2024coohoi} adapts learned interaction policies to cooperative manipulation through object-level coupling. Such transfer is typically realized through policy initialization or task-specific fine-tuning. Although the adapted policy can solve a specific cooperative task, it still entangles low-level motor skill and high-level coordination within a single task policy. As a result, the learned behavior is difficult to reuse, compose, or transfer to new object geometries and cooperative configurations.

Our key insight is that a single-human HOI policy should not be directly fine-tuned into a task-specific multi-human policy. Instead, it should be preserved and converted into a reusable \emph{Object-oriented Motion Skill} that can realize local proxy motions across downstream HOI tasks. With this skill, downstream learning only needs to reason in the \emph{Object-oriented Action Space}. This leads to a hierarchical and decoupled control framework, where the low-level Object-oriented Motion Skill handles contact-rich execution and the high-level policy performs compositional object-level coordination through proxy-motion planning.

To make the skill reusable, the control interface should neither be tied to a specific task goal nor over-constrain the humanoid motion. We therefore define the Object-oriented Action Space as short-horizon motions of an object proxy, allowing the low-level policy to learn how to realize desired proxy motion through contact selection, balance control, and full-body coordination.

Specifically, we reinterpret rollouts from single-human HOI teacher policies as object-oriented action supervision by extracting future object-proxy motions from their executed trajectories. This converts task-specific teachers into an Object-oriented Motion Skill that can be frozen and reused as a low-level executor. In downstream tasks, a high-level policy observes the shared object, task goal, agent states, and local manipulation regions, and generates region-wise object-oriented actions. In this way, downstream RL no longer optimizes contact-rich full-body control directly, but instead learns cooperative decision making within the compact Object-oriented Action Space.

Our contributions are threefold:
\begin{itemize}[leftmargin=*]
\item We establish a transferable formulation for physics-based multi-human HOI, shifting the focus from task-specific cooperative policy learning to transferable object-oriented motion coordination.

\item We develop a distillation-based hierarchical learning framework that converts single-human HOI knowledge into a reusable Object-oriented Motion Skill, enabling high-level policies to coordinate multiple humans through short-horizon object-proxy motions.

\item We validate the framework on cooperative HOI tasks, showing that an Object-oriented Motion Skill distilled from a single-human skill supports robust proxy-motion execution and enables high-level policy learning across different interaction types, object geometries, and team sizes.
\end{itemize}

%% file: sec/relatedwork.tex
\section{Related Work}
\label{sec:relatedwork}

\begin{figure*}[t]
    \centering
    \includegraphics[width=0.8\textwidth]{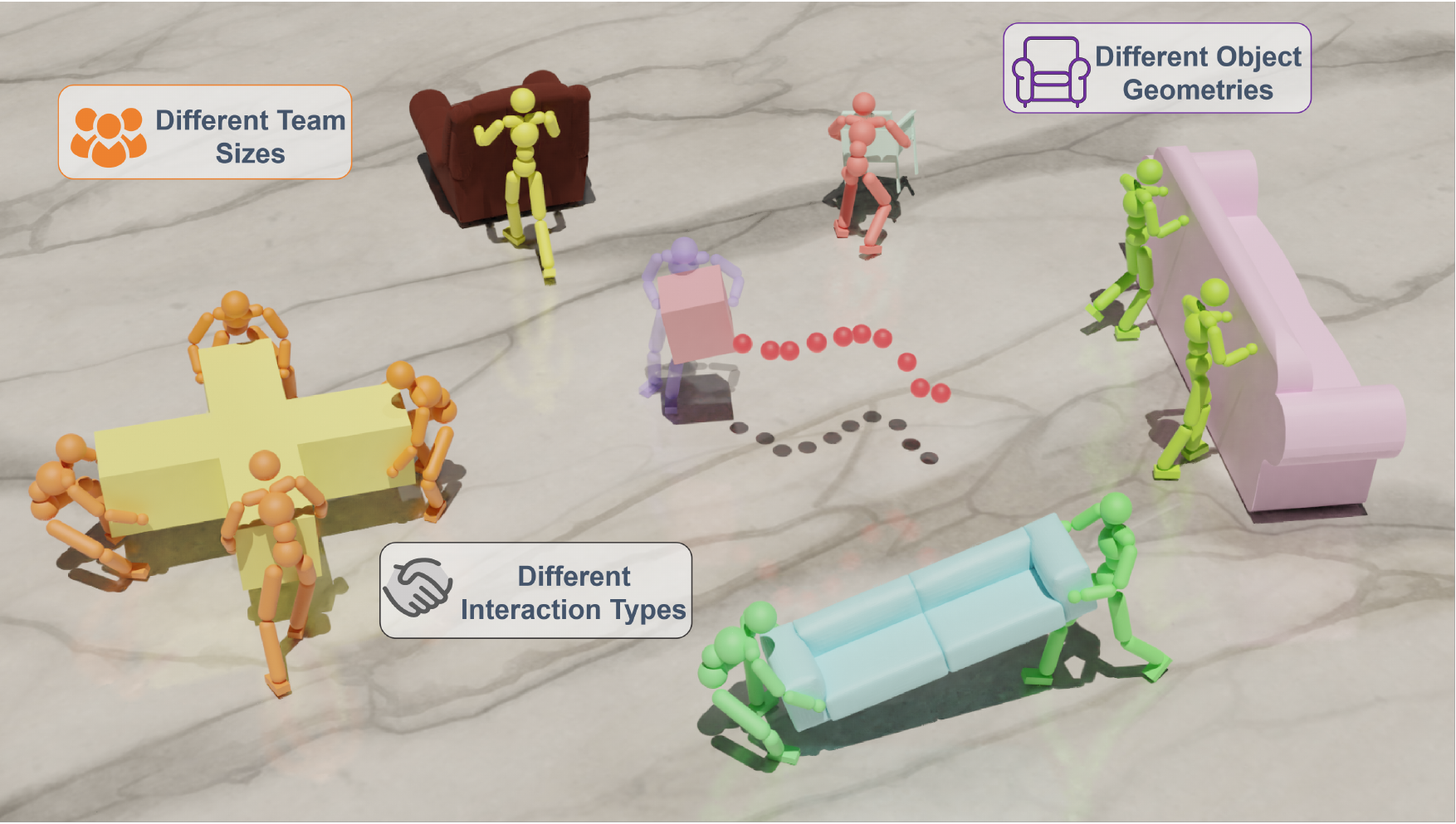}
    \caption{
Our method composes object-oriented motion skills and cooperative planning to support shared object manipulation across team sizes, object geometries, and interaction types.
    }
    \label{fig:teaser}
\end{figure*}

\paragraph{Human-Object Interaction Motion Synthesis and Physics-Based Control.}
Character animation and motion synthesis have extensively studied human motion in 3D scenes and human-object interaction, including kinematic human-scene and dynamic-object interaction methods~\citep{li2019putting,zhang2020generating,Xuan_2023_ICCV,Zhao:ECCV:2022,lee2023lama,wang2022humanise,wang2024move,starke2019neural,Zhao:ICCV:2023,starke2020local,jiang2022chairs,zhang2022couch,zhang2024core4d}. 
Recent diffusion-based approaches further improve motion diversity and controllability by conditioning on scene constraints, text, action labels, or object trajectories~\citep{ho2020denoising,huang2023diffusion,tang2024unified,yi2025generating,zhao2024dart,li2023object,li2023controllable,pi2023hierarchical,peng2023hoi,zhangopenhoi,jiang2024scaling,jiang2024autonomous}. 
However, such kinematic generation methods often lack physical validity, especially for long-horizon contact-rich interactions that require sustained contacts, force exchange, and object dynamics.

Physics-based character control addresses this issue by learning closed-loop policies for physically plausible full-body behaviors from motion imitation, adversarial motion priors, latent skills, and large-scale motion datasets~\citep{peng2018deepmimic,peng2021amp,peng2022ase,tessler2023calm,luo2024universal,juravsky2024superpadl,tessler2024maskedmimic}. 
These techniques have been extended to physics-based HOI and human-scene interaction, enabling contact-rich behaviors such as object catching and manipulation, scene-level interactions, dynamic HOI imitation, reusable sports skills, task-tokenized scene interaction, and whole-body HOI control across diverse objects~\citep{merel2020catch,chao2021learning,xie2023hierarchical,hassan2023synthesizing,wang2023physhoi,wang2024skillmimic,zhang2023learning,liu2017learning,liu2022motor,xie2022learning,pan2025tokenhsi,xu2025intermimic}. 
While these methods substantially improve physical realism and task coverage, their control abstractions are usually tied to reference motions, task labels, or single-human interaction settings. 

\paragraph{Multi-Character Interaction and Cooperative Object Manipulation.}
Compared with single-human HOI, cooperative interaction introduces a stronger coordination challenge: a shared object-level task must be decomposed into per-human contacts, motions, and force application. 
Existing multi-person datasets and kinematic animation methods have enabled visually plausible human-human interactions and choreographic motions~\citep{kundu2020cross,shen2019interaction,fieraru2020three,xu2024inter,liu2025core4d,zhang2024hoi,kogashi2026mmhoi,le2023music,siyao2024duolando,won2014generating,liang2024intergen,tanaka2023role,shafir2023human,ghosh2024remos,tan2025think,zhang2025reactffusion,zhao2025freedance,fan2024freemotion,ghosh2025duetgen,javed2024intermask}. 
However, these methods mainly target motion synthesis and do not guarantee physically valid contacts, forces, or shared object dynamics. 
Recent physics-based multi-character methods improve physical plausibility by combining generated motions with controllers or converting multi-human motions into executable policies~\citep{huang2025diffuse,tevet2024closd,karunratanakul2023guided,zhang2024tedi,luo2023perpetual,liu2024physreaction,ji2025towards,li2025interagent}, but they mostly focus on human-human interaction, sports, reaction, or collision avoidance rather than contact-rich cooperative object manipulation~\citep{haworth2020deep,rempe2023trace,luo2024smplolympics}. 

Cooperative object manipulation has also been explored in legged robotics, including multi-robot payload transport, pushing, and collaborative carrying~\citep{nachum2020multiagent,devincenti2023centralized,10341700,Turrisi_2024,feng2025learning,pandit2025multiquadrupedcooperativeobjecttransport}. 
These systems demonstrate object-level coordination, but they typically rely on quadrupedal morphology, engineered manipulation interfaces, or task-specific control designs, and do not address full-body humanoid HOI with motion priors. Most closely related to our setting, CooHOI~\citep{gao2024coohoi} learns cooperative human-object interaction through shared object dynamics, but its adaptation still couples low-level motor execution with task-specific multi-human coordination. In contrast, our method preserves single-human HOI skills as a reusable low-level skill operating over object-oriented proxy motions, enabling high-level policies to compose local object motions across object geometries, interaction types, and team sizes.

%% file: sec/method.tex
\section{Method}
\label{sec:method}

\begin{figure*}[t]
    \centering
    \includegraphics[width=\textwidth]{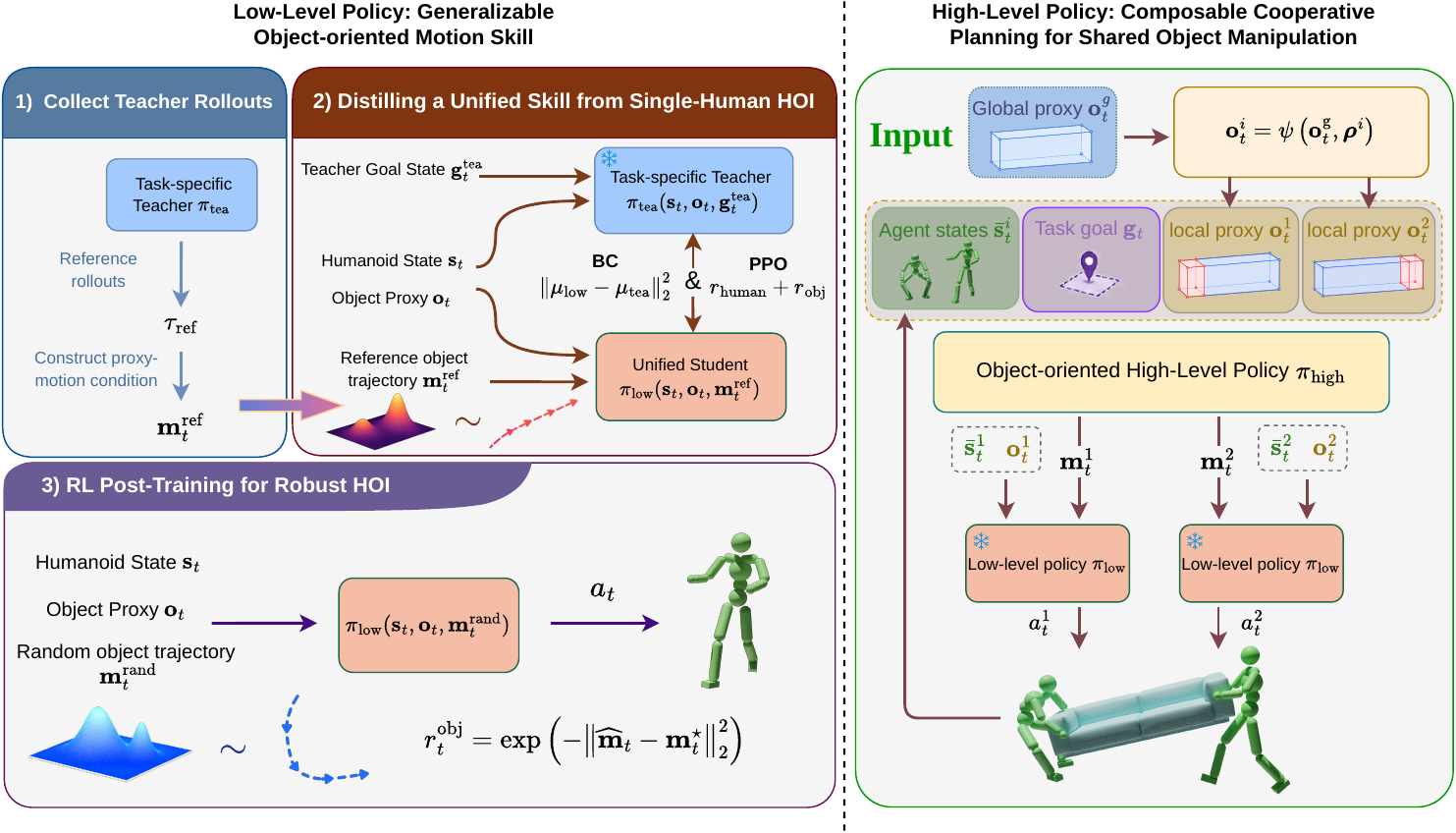}
    \caption{
Overview of our hierarchical framework. We distill and post-train a reusable Object-oriented Motion Skill from task-specific single-human HOI rollouts re-labeled as proxy-motion supervision. The frozen skill then executes local proxy motions planned by high-level policies, enabling composable single- and multi-human HOI.
}
    \label{fig:pipeline}
\end{figure*}

Our method builds a composable HOI control hierarchy around object-oriented proxy motions.
Sec.~\ref{sec:object_motion_skill} first introduces the \emph{Object-oriented Motion Skill}, a reusable low-level policy that realizes short-horizon motions of object proxies through physically feasible humanoid control.
Its \emph{Object-oriented Action Space} provides a compact interface between high-level task reasoning and contact-rich execution.
Sec.~\ref{sec:object_centric_hierarchy} then trains task-specific high-level policies on top of the frozen skill, allowing single- and multi-human HOI tasks to be solved by generating region-wise proxy motions rather than full-body motions or torque-level actions.
Fig.~\ref{fig:pipeline} provides an overview of the full pipeline.

\subsection{Low-Level Policy: Generalizable Object-oriented Motion Skill}
\label{sec:object_motion_skill}

\paragraph{Definition of Object-oriented Action Space.}
We decouple high-level HOI reasoning from low-level physical execution by introducing an
\emph{Object-oriented Motion Skill}, a reusable low-level policy controlled through an
\emph{Object-oriented Action Space}.
Instead of directly tracking full-body reference motions or optimizing toward a task-specific goal, the skill realizes short-horizon motions of a manipulated object proxy.
For clarity, we formulate the action space for a single humanoid and omit the agent index.

Let $\mathbf{s}_t$ denote the humanoid state and $\mathbf{o}_t$ denote the object proxy observed by the low-level skill.
We represent either a complete object or a task-relevant local region as a compact cuboid proxy with unified kinematic and bounding-box features.
This abstraction provides a shared input format across manipulation regions, object geometries, and interaction types.

We define an object anchor $\mathbf{p}_t \in \mathbb{R}^3$ as the reference point of the proxy $\mathbf{o}_t$, such as its root or center.
The Object-oriented Action Space is defined as a short-horizon sequence of anchor displacements:
\begin{equation}
    \mathbf{m}_t
    =
    \left[
    \Delta \mathbf{p}_{t,0},
    \Delta \mathbf{p}_{t,1},
    \ldots,
    \Delta \mathbf{p}_{t,K-1}
    \right],
    \qquad
    \Delta \mathbf{p}_{t,k}
    =
    \mathbf{p}_{t+k+1}
    -
    \mathbf{p}_{t+k},
    \label{eq:object_oriented_action}
\end{equation}
where $K$ is the prediction horizon.
Eq.~\eqref{eq:object_oriented_action} is written in a common coordinate frame for notation.
In practice, $\mathbf{m}_t$ is transformed into the local frame of the receiving humanoid before being fed to the skill.
Thus, $\mathbf{m}_t$ specifies the desired near-future motion of the object proxy, while $\mathbf{o}_t$ provides the geometric and dynamic context required for physical execution.

Given the humanoid state, the object proxy, and the object-oriented action, the low-level skill outputs humanoid control actions:
\begin{equation}
    \mathbf{a}_t
    =
    \pi_{\mathrm{low}}
    \left(
    \mathbf{s}_t,
    \mathbf{o}_t,
    \mathbf{m}_t
    \right).
    \label{eq:ooms_policy}
\end{equation}
Here, $\mathbf{m}_t$ is the high-level object-oriented action, whereas $\mathbf{a}_t$ is the low-level humanoid control action.
Compared with a final object goal, $\mathbf{m}_t$ provides dense temporal guidance.
Compared with full-body motion tracking, it preserves freedom in contact selection, balance, and locomotion.
This interface allows downstream high-level policies to compose diverse single- and multi-human HOI tasks by generating proxy-level object motions while reusing the same low-level skill.

\paragraph{Trajectory Rollouts from Task-specific Teacher Policy.}
Existing single-human HOI policies are typically trained with task-specific goals, hand-designed rewards, and AMP-based motion priors~\citep{peng2021amp,hassan2023synthesizing}.
We treat such a trained policy as a teacher, denoted by
$\mathbf{a}^{\mathrm{tea}}_t \sim \pi_{\mathrm{tea}}(\cdot \mid \mathbf{s}_t,\mathbf{o}_t,\mathbf{g}^{\mathrm{tea}}_t)$,
where $\mathbf{s}_t$, $\mathbf{o}_t$, and $\mathbf{g}^{\mathrm{tea}}_t$ are the humanoid state, object proxy state, and task-specific teacher goal, respectively.
Although $\mathbf{g}^{\mathrm{tea}}_t$ is effective for the source task, it is tied to a particular goal and reward formulation and therefore cannot directly serve as a reusable low-level control interface.

We collect successful teacher rollouts and convert them into supervision for the Object-oriented Motion Skill.
A reference rollout is written as
$\tau_{\mathrm{ref}}=\{(\mathbf{s}^{\mathrm{ref}}_t,\mathbf{o}^{\mathrm{ref}}_t,\mathbf{p}^{\mathrm{ref}}_t,\mathbf{g}^{\mathrm{tea}}_t)\}_{t=0}^{H-1}$,
where $\mathbf{p}^{\mathrm{ref}}_t$ denotes the anchor position of the object proxy.
These rollouts provide physically plausible interaction states, while their object proxy trajectories define training conditions in the Object-oriented Action Space.
For each reference state, we construct the future proxy-motion condition
$\mathbf{m}^{\mathrm{ref}}_t=[\Delta\mathbf{p}^{\mathrm{ref}}_{t,0},\ldots,\Delta\mathbf{p}^{\mathrm{ref}}_{t,K-1}]$,
with
$\Delta\mathbf{p}^{\mathrm{ref}}_{t,k}=\mathbf{p}^{\mathrm{ref}}_{t+k+1}-\mathbf{p}^{\mathrm{ref}}_{t+k}$,
following Eq.~\eqref{eq:object_oriented_action}.
The original teacher goal is retained only for querying $\pi_{\mathrm{tea}}$, whereas the student is conditioned on the constructed future object motion:
the teacher uses $\pi_{\mathrm{tea}}(\mathbf{a}_t \mid \mathbf{s}_t,\mathbf{o}_t,\mathbf{g}^{\mathrm{tea}}_t)$,
while the student uses $\pi_{\mathrm{low}}(\mathbf{a}_t \mid \mathbf{s}_t,\mathbf{o}_t,\mathbf{m}^{\mathrm{ref}}_t)$.
In this way, a task-specific single-human HOI policy is reinterpreted as object-oriented action supervision for a reusable low-level skill.
Teacher actions are not stored in the rollout buffer; instead, they are queried online when behavior-cloning supervision is applied.

\paragraph{Distilling a Unified Skill from Single-Human HOI.}
In the first stage, we use the converted teacher rollouts to bootstrap the Object-oriented Motion Skill from physically feasible contact-rich interactions.
Each environment is initialized from a sampled reference frame, and the executed action is replaced by the online teacher action with probability $p$, which is gradually annealed to zero.
This defines a mixed rollout policy $\pi_{\mathrm{mix}}^{(p)}$ that exposes the student to feasible interaction states during early training.
The student is trained with PPO conditioned on the future proxy-motion condition, together with an auxiliary online behavior cloning loss:
\begin{equation}
    \mathcal{L}
    =
    -
    \mathbb{E}_{\tau \sim \pi_{\mathrm{mix}}^{(p)}}
    \left[
    \sum_{t=0}^{H-1}
    \gamma^t r_t
    \right]
    +
    \lambda_{\mathrm{BC}}
    \mathbb{E}_{t}
    \left[
    \left\|
    \mu_{\mathrm{low}}(\mathbf{s}_t,\mathbf{o}_t,\mathbf{m}^{\mathrm{ref}}_t)
    -
    \mu_{\mathrm{tea}}(\mathbf{s}_t,\mathbf{o}_t,\mathbf{g}^{\mathrm{tea}}_t)
    \right\|_2^2
    \right],
    \label{eq:oo_distill_objective}
\end{equation}
where $\mu_{\mathrm{low}}$ and $\mu_{\mathrm{tea}}$ denote the mean actions of the student and teacher policies.
Here, $r_t$ combines an object-proxy trajectory tracking reward $r_t^\mathrm{obj}$ and an auxiliary human pose tracking reward $r_t^\mathrm{human}$.
The object reward trains the skill to follow the future proxy-motion condition, while the human reward stabilizes early contact-rich imitation by matching the reference motion.
We provide the details of the human pose tracking reward in the appendix.
Both $p$ and $\lambda_{\mathrm{BC}}$ are decayed during training, so that the student gradually shifts from teacher-guided imitation to reinforcement learning under the Object-oriented Action Space.

The core reward is object-proxy trajectory tracking.
For a target proxy-motion condition
$\mathbf{m}^{\star}_t=[\Delta\mathbf{p}^{\star}_{t,0},\ldots,\Delta\mathbf{p}^{\star}_{t,K-1}]$,
we define
\begin{equation}
    r_t^\mathrm{obj}
    =
    \exp
    \left(
    -
    \frac{1}{K}
    \sum_{k=0}^{K-1}
    \left\|
    \widehat{\Delta\mathbf{p}}_{t,k}
    -
    \Delta\mathbf{p}^{\star}_{t,k}
    \right\|_2^2
    \right),
    \label{eq:object_tracking_reward}
\end{equation}
where $\widehat{\Delta\mathbf{p}}_{t,k}$ denotes the executed anchor displacement.
In this stage, $\mathbf{m}^{\star}_t=\mathbf{m}^{\mathrm{ref}}_t$.
The reference human motion is used only for reward computation and is never provided to the student input.
This prevents the student from becoming a full-body motion tracker and forces it to realize the demonstrated interaction through object-oriented proxy-motion conditions.

\paragraph{RL Post-Training for Robust HOI.}
The distillation stage trains the policy mainly within the trajectory distribution induced by successful teacher rollouts. However, in downstream HOI tasks, high-level policies may generate proxy motions that deviate from the teacher demonstrations, causing the low-level policy to encounter out-of-distribution states.
To improve the robustness and out-of-distribution generalization of the Object-oriented Motion Skill, we further post-train $\pi_{\mathrm{low}}$ with randomly generated feasible object-proxy motions. In this stage, we remove teacher action replacement and the behavior cloning loss, i.e., $p=0$ and $\lambda_{\mathrm{BC}}=0$, and optimize the policy only with the object-proxy tracking reward in Eq.~\eqref{eq:object_tracking_reward}.
The target motion is replaced by a sampled random proxy-motion condition,
$\mathbf{m}^{\star}_t=\mathbf{m}^{\mathrm{rand}}_t$.
This exposes the policy to a broader range of object motions and forces it to realize them through physical humanoid-object interaction rather than relying on teacher-guided trajectories.
After post-training, $\pi_{\mathrm{low}}$ is frozen and reused as the shared Object-oriented Motion Skill for downstream single- and multi-human HOI policy learning.

\subsection{High-Level Policy: Composable Cooperative Planning for Shared Object Manipulation}
\label{sec:object_centric_hierarchy}

The learned Object-oriented Motion Skill enables downstream HOI tasks to be formulated as proxy-motion planning over task-relevant manipulation regions.
Instead of generating full-body humanoid motions or low-level control actions, the high-level policy predicts object-oriented actions that specify how local object proxies should move in the near future.
The frozen Object-oriented Motion Skill then realizes these proxy-level motions through physical humanoid control.
This forms a composable hierarchy: downstream tasks only train task-specific high-level policies, while sharing the same low-level Object-oriented Motion Skill.

For a task with $N$ humanoids, we assign each humanoid a manipulation region $\boldsymbol{\rho}^{i}$ according to the object geometry and interaction role.
This region specifies where the humanoid interacts with the object, but does not prescribe its full-body motion, contact pattern, or the complete object transport strategy.
Given the global object proxy $\mathbf{o}^{\mathrm{g}}_t$, we instantiate a compatible local proxy by
\begin{equation}
    \mathbf{o}^{i}_t
    =
    \psi
    \left(
    \mathbf{o}^{\mathrm{g}}_t,
    \boldsymbol{\rho}^{i}
    \right),
    \qquad
    i=1,\ldots,N .
    \label{eq:local_proxy}
\end{equation}
Here, $\psi$ is a geometric adapter that represents the selected region using the same cuboid proxy format as in Object-oriented Motion Skill learning.
A local proxy may correspond to an endpoint, side, or corner of a larger object.
Since $\mathbf{o}^{i}_t$ remains compatible with the skill observation space, the same Object-oriented Motion Skill can be applied to different object regions, geometries, interaction types, and team sizes without architectural changes.

The high-level policy adopts an object-centric view and outputs one object-oriented action for each local proxy:
\begin{equation}
    \left\{
    \mathbf{m}^{i}_t
    \right\}_{i=1}^{N}
    =
    \pi_{\mathrm{high}}
    \left(
    \mathbf{o}^{\mathrm{g}}_t,
    \left\{
    \bar{\mathbf{s}}^{i}_t,
    \mathbf{o}^{i}_t
    \right\}_{i=1}^{N},
    \mathbf{g}_t
    \right),
    \qquad
    \mathbf{a}^{i}_t
    =
    \pi_{\mathrm{low}}
    \left(
    \mathbf{s}^{i}_t,
    \mathbf{o}^{i}_t,
    \mathbf{m}^{i}_t
    \right),
    \label{eq:hierarchical_execution}
\end{equation}
where $\bar{\mathbf{s}}^{i}_t$ denotes the state of humanoid $i$ represented in the object-centric frame, and $\mathbf{g}_t$ is the task goal state.
Each $\mathbf{m}^{i}_t$ follows the Object-oriented Action Space defined in Eq.~\eqref{eq:object_oriented_action}, but specifies the desired near-future motion of the corresponding local proxy.
Humanoids are coordinated by the high-level policy, the shared object dynamics, and object-level task rewards, while the frozen Object-oriented Motion Skill handles contact-rich full-body execution.

During downstream training, $\pi_{\mathrm{low}}$ is fixed and only $\pi_{\mathrm{high}}$ is optimized.
Following prior physics-based interaction-control methods~\citep{hassan2023synthesizing, gao2024coohoi}, we optimize $\pi_{\mathrm{high}}$ with PPO using task-specific rewards; details are provided in the appendix.
Thus, downstream HOI learning is reduced to region-conditioned proxy-motion planning, enabling the same Object-oriented Motion Skill to support single-human and multi-human HOI across different object geometries and interaction types.

%% file: sec/experiment.tex
\section{Experiments}
\label{sec:experiments}

\subsection{Evaluation on Compositional HOI Tasks}
\label{sec:exp_transfer}

\paragraph{Evaluation Protocol.}
We evaluate whether the Object-oriented Motion Skill learned from single-human interaction can serve as a reusable low-level interface for downstream object manipulation. In all tasks, $\pi_{\mathrm{low}}$ is frozen, and only a task-specific $\pi_{\mathrm{high}}$ is trained to predict object-oriented actions for local proxies, which are then executed by the frozen skill. We test multi-human carrying and pushing, including two-human sofa carrying, four-human H-box/cross-box carrying, and two-/four-human sofa pushing. We also include single-human armchair/table carrying and large-box pushing to evaluate transfer across object geometries and interaction modes. Each object or manipulation region is represented by the same cuboid proxy format as in skill learning, allowing all settings to share the Object-oriented Action Space without modifying $\pi_{\mathrm{low}}$.All experiments are conducted in Isaac Gym~\citep{makoviychuk2021isaac} with 4096 parallel environments. We use the same humanoid model as prior physics-based HOI controllers~\citep{hassan2023synthesizing,gao2024coohoi}, with 15 rigid bodies and 28 PD-controlled joints. Objects are initialized with random yaw in $[0,2\pi]$ and root-to-target distance of $4$--$10\,\mathrm{m}$. We report success rate(\%) and precision(cm) over 4096 environments, with success defined as the object root reaching within $0.2\,\mathrm{m}$ of the target.

\paragraph{Baselines.}
We compare with CooHOI~\citep{gao2024coohoi}, a representative physics-based cooperative HOI baseline.
CooHOI adapts a full interaction policy to each downstream carrying task, but still couples motor execution and coordination in one policy, limiting direct reuse across geometries, interaction modes, and team sizes. Our method instead freezes the Object-oriented Motion Skill and trains only a task-specific high-level policy in the Object-oriented Action Space. We evaluate CooHOI on its supported carrying tasks; for pushing, which is outside its released carrying-focused setting, we report only our method to test generalization beyond source carrying.

\input{table/hrl}

\paragraph{Compositional Transfer.}
Table~\ref{tab:hrl} shows that our hierarchy transfers across object geometries, interaction modes, and team sizes.
For multi-human carrying, the method remains effective from two-human sofa carrying to four-human compositional-object manipulation, since the high-level policy coordinates local proxy motions while the frozen Object-oriented Motion Skill handles contact-rich execution. This avoids binding the policy to a specific cooperative layout as in monolithic full-body adaptation. The pushing results further verify interaction-mode transfer: although the skill is distilled from single-human carrying, it supports two- and four-human sofa pushing by changing only the high-level policy. Single-human armchair/table carrying and large-box pushing additionally confirm geometric transfer, showing that the cuboid-proxy abstraction provides a unified and reusable low-level interface across object shapes and tasks.

\subsection{Evaluation on Object-oriented Motion Skill}
\label{sec:exp_distill}

\paragraph{Evaluation Protocol.}
We evaluate whether the proposed pipeline can convert a task-specific single-human HOI policy into a reusable Object-oriented Motion Skill. The student is conditioned on the humanoid state, object-proxy observation, and short-horizon object-oriented action. We test two target distributions: \emph{teacher-rollout targets}, relabeled from successful teacher trajectories, and \emph{random proxy-motion targets}, procedurally generated to evaluate generalization beyond the teacher-induced motion manifold. All experiments use the simulator and humanoid model of Sec.~\ref{sec:exp_transfer}. The teacher policy is trained independently on the source single-human carrying task in Sec.~\ref{sec:object_motion_skill}. During skill learning, the box size is randomized within $[0.5,1.5]$ times its nominal scale to improve geometric coverage. We report tracking errors on the object-proxy anchor, i.e., the box center in our implementation. $E_{\mathrm{pos}}$, $E_{\mathrm{vel}}$, and $E_{\mathrm{acc}}$ denote mean position(mm), velocity(mm/frame), and acceleration errors(mm/frame$^2$). Trajectory Targets Reached (TTR) is the fraction of target positions reached within $15\,\mathrm{cm}$. Succ$_{\mathrm{traj}}$ measures complete trajectory-following success, where failure occurs if the object deviates more than $30\,\mathrm{cm}$ from the reference at any timestep.

\paragraph{Baselines.}
We compare with adapted variants of two representative physics-based HOI skill-learning paradigms.
InterPhys~\citep{hassan2023synthesizing} represents goal-driven interaction control, where policies are trained with task goals, task rewards, and adversarial motion priors constructed from OMOMO~\citep{li2023controllable} and AMASS~\citep{AMASS:ICCV:2019} motion clips.
SkillMimic~\citep{wang2024skillmimic} represents imitation-based skill learning, where policies track human-object demonstrations conditioned on skill labels.
For a controlled comparison, we adapt both methods to receive the same short-horizon object-oriented action and optimize them with the same object-proxy tracking objective.
Thus, all methods share the same target motion interface, while differing in how the underlying contact-rich interaction skill is acquired.

\input{table/distill}

\paragraph{Proxy-motion Tracking.}
Table~\ref{tab:distill} shows that our method achieves the best tracking performance on teacher-rollout targets.
This indicates that relabeling successful teacher rollouts into future proxy motions preserves physically feasible interaction behavior without exposing the student to full-body reference motion.
The advantage becomes more pronounced on random proxy-motion targets.
Although the adapted baselines receive the same object-oriented action, their performance drops when the target motions move away from the teacher-rollout distribution.
In contrast, our Object-oriented Motion Skill maintains higher target-reaching and trajectory success rates, showing that teacher-guided relabeling and random-motion RL post-training expand the executable region of the Object-oriented Action Space.
These results support object-oriented motion as a compact low-level interface: it provides sufficient temporal guidance for object transport while preserving flexibility in contact selection, balance, and locomotion.

\paragraph{Ablation.}
Table~\ref{tab:distill} further validates the contribution of the two training stages.
Removing online distillation causes a clear performance drop, especially on random proxy-motion targets.
This suggests that object-proxy tracking rewards alone are insufficient to bootstrap contact-rich carrying, since stable grasping, balance, and locomotion must first emerge before trajectory rewards become informative.
Teacher-guided rollout and online behavior cloning therefore provide a feasible-interaction prior that stabilizes early skill acquisition.

Removing random-motion RL post-training preserves strong performance on teacher-rollout targets but substantially weakens generalization to random proxy-motion targets.
This indicates that relabeled teacher rollouts provide high-quality supervision near the source task manifold, but cover only a limited range of executable object motions.
Random-motion post-training expands this range and turns the distilled policy from a teacher-rollout follower into a reusable Object-oriented Motion Skill for downstream high-level policy learning.

%% file: table/hrl.tex
\begin{table}[t]
\centering
\small
\setlength{\tabcolsep}{4pt}
\caption{
Quantitative evaluation of downstream compositional HOI tasks. The Object-oriented Motion Skill is frozen across all tasks, and only task-specific high-level policies are trained to generate object-oriented actions for local proxies. 
}
\label{tab:hrl}
\begin{tabular}{clc cc cc}
\toprule
\multirow{2}{*}{Task}
& \multirow{2}{*}{Object}
& \multirow{2}{*}{\#Agents}
& \multicolumn{2}{c}{CooHOI}
& \multicolumn{2}{c}{Ours} \\
\cmidrule(lr){4-5} \cmidrule(lr){6-7}
& &
& Succ. $\uparrow$
& Precision $\downarrow$
& Succ. $\uparrow$
& Precision $\downarrow$ \\
\midrule
\multirow{5}{*}{\textit{Carrying}}
& Armchair  & 1 & 97.26 & 5.05  & \textbf{97.81} & \textbf{4.99} \\
& Table     & 1 & 97.07 & 5.23  & \textbf{97.19} & \textbf{5.11} \\
& Sofa      & 2 & 84.17 & 10.12 & \textbf{88.21} & \textbf{8.70} \\
& H-box     & 4 & 80.96 & 17.50 & \textbf{85.63} & \textbf{14.24} \\
& Cross-box & 4 & 82.38 & 15.13 & \textbf{86.14} & \textbf{13.81} \\
\midrule
\multirow{3}{*}{\textit{Pushing}}
& Large box & 1 & --- & --- & 96.24 & 9.31 \\
& Sofa      & 2 & --- & --- & 94.98 & 15.22 \\
& Sofa      & 4 & --- & --- & 88.14 & 17.97 \\
\bottomrule
\end{tabular}
\end{table}

%% file: table/distill.tex
\begin{table*}[t]
    \centering
    \caption{
    Quantitative evaluation of Object-oriented Motion Skill learning. We report proxy-motion tracking under teacher-rollout targets and procedurally generated random proxy-motion targets.
    }
    \label{tab:distill}
    \resizebox{\textwidth}{!}{
    \begin{tabular}{lcccccccccc}
        \toprule
        \multirow{2}{*}{Method}
        & \multicolumn{5}{c}{Teacher-Rollout Targets}
        & \multicolumn{5}{c}{Random-Motion Targets} \\
        \cmidrule(lr){2-6} \cmidrule(lr){7-11}
        & $E_{\mathrm{pos}}$ $\downarrow$
        & $E_{\mathrm{vel}}$ $\downarrow$
        & $E_{\mathrm{acc}}$ $\downarrow$
        & TTR $\uparrow$
        & Succ$_{\mathrm{traj}}$ $\uparrow$
        & $E_{\mathrm{pos}}$ $\downarrow$
        & $E_{\mathrm{vel}}$ $\downarrow$
        & $E_{\mathrm{acc}}$ $\downarrow$
        & TTR $\uparrow$
        & Succ$_{\mathrm{traj}}$ $\uparrow$ \\
        \midrule
        InterPhys~\citep{hassan2023synthesizing}
        & 106.5 & 47.8 & 50.9 & 82.6 & 68.2
        & 112.2 & 63.9 & 54.1 & 76.4 & 60.8 \\
        SkillMimic~\citep{wang2024skillmimic}
        & 91.0 & 40.3 & 44.2 & 90.8 & 81.2
        & 99.3 & 54.6 & 49.7 & 84.9 & 68.3 \\
        \midrule
        \textbf{Ours}
        & \textbf{79.8} & \textbf{38.2} & \textbf{31.3} & \textbf{100.0} & \textbf{98.9}
        & \textbf{95.8} & \textbf{45.2} & \textbf{39.2} & \textbf{96.3} & \textbf{91.8} \\
        Ours w/o Online Distill.
        & 117.9 & 52.3 & 50.8 & 86.7 & 84.8
        & 148.2 & 63.1 & 55.9 & 58.4 & 23.8 \\
        Ours w/o Random-Motion RL
        & 88.1 & 41.3 & 32.8 & 97.2 & 98.1
        & 156.1 & 65.5 & 57.2 & 63.1 & 38.7 \\
        \bottomrule
    \end{tabular}}
\end{table*}

%% file: sec/limitation.tex
\section{Limitations}
Our framework studies compositional HOI within a reusable Object-oriented Motion Skill, where high-level policies coordinate local proxy motions through a shared low-level executor. 
However, it does not explicitly address composition across heterogeneous skill types or long-horizon temporal task structures, such as multi-stage task decomposition and dynamic role reassignment. 
Incorporating skill-level composition with higher-level planning is a promising direction for future work.

%% file: sec/conclusion.tex
\section{Conclusion}
We presented a hierarchical framework for composable physics-based HOI that converts task-specific single-human interaction policies into a reusable \emph{Object-oriented Motion Skill}. By relabeling teacher rollouts with short-horizon object-proxy motions, our method learns a low-level skill under the \emph{Object-oriented Action Space}, instead of directly fine-tuning single-human policies into task-specific cooperative controllers. The frozen skill realizes region-wise object-oriented actions through contact-rich full-body control, allowing downstream policies to transfer across different object geometries, interaction types, and team sizes.

%% file: sec/appendix.tex
\section{Appendix}

\subsection{Training and Implementation Details}
\label{appendix:training}

\paragraph{Character Model and Policy Inputs.}
We use a customized simulated humanoid adapted from the AMP character model~\cite{peng2021amp}.
The humanoid consists of $15$ rigid bodies, $12$ actuated joints, and $28$ degrees of freedom (DoFs).
The $28$ DoFs correspond to the actuated joint space controlled by low-level PD servos, where the policy predicts target joint configurations rather than directly applying torques.
At each time step $t$, the proprioceptive state $\mathbf{s}_t$ describes the physical state of the character.
Following AMP~\cite{peng2021amp}, it includes the positions, rotations, linear velocities, and angular velocities of all rigid bodies.
All quantities are expressed in the character-centric local frame, except for the root orientation and root height, which are kept in the world frame to preserve global elevation information.
We represent body rotations using the $6$D rotation representation and exclude the global root position from the proprioceptive input.
This yields a $223$D state vector $\mathbf{s}_t \in \mathbb{R}^{223}$.
At each time step, the policy outputs an action $\mathbf{a}_t$ that specifies target joint rotations for PD control.
For $3$D spherical joints, the target rotations are parameterized as exponential maps.
No external perturbation forces are applied to the humanoid during simulation.

For low-level skill learning, we formulate proxy tracking as a short-horizon object-motion execution task.
The controlled humanoid interacts with a cuboid object proxy, whose anchor $\mathbf{p}_t\in\mathbb{R}^{3}$ is defined as the proxy center.
The object proxy observation $\mathbf{o}_t \in \mathbb{R}^{54}$ encodes the geometric and kinematic state of the oriented cuboid proxy, including its vertices $\mathbf{V}^{\mathrm{box}}_t \in \mathbb{R}^{8 \times 3}$, orientation $\boldsymbol{\theta}_t \in \mathbb{R}^{3}$, vertex linear velocities $\mathbf{v}^{\mathrm{box}}_t \in \mathbb{R}^{8 \times 3}$, and angular velocity $\boldsymbol{\omega}_t \in \mathbb{R}^{3}$.
Following the Object-oriented Action Space in Eq.~\eqref{eq:object_oriented_action}, the task goal is a future proxy-motion condition
\begin{equation}
    \mathbf{m}^{\star}_t
    =
    [\Delta\mathbf{p}^{\star}_{t,0},\ldots,\Delta\mathbf{p}^{\star}_{t,K-1}],
\end{equation}
where we use $K=10$, resulting in $\mathbf{m}^{\star}_t\in\mathbb{R}^{30}$.

Proxy-motion targets may be constructed in different reference frames, such as the teacher rollout frame, the object-centric frame, or the global simulation frame.
Before being provided to the low-level policy, they are transformed into the receiving humanoid's local frame, following the object-motion interface in Sec.~\ref{sec:object_motion_skill}.
For simplicity, $\mathbf{m}^{\star}_t$ denotes this frame-consistent proxy-motion condition.
For notation simplicity, we use $\mathbf{m}^{\star}_t$ to denote the frame-consistent proxy-motion condition provided to $\pi_{\mathrm{low}}$.
Together with the humanoid proprioceptive state and proxy observation, the low-level policy acts as
\begin{equation}
    \mathbf{a}_t
    =
    \pi_{\mathrm{low}}
    \left(
    \mathbf{s}_t,\mathbf{o}_t,\mathbf{m}^{\star}_t
    \right).
\end{equation}

\paragraph{Low-Level Skill Training.}
The Object-oriented Motion Skill is trained in two consecutive stages, as summarized in Algorithm~\ref{alg:low_level_training}.
The first stage distills a task-specific single-human teacher into a proxy-motion-conditioned student.
Given successful teacher rollouts, we relabel each reference state with a future object-oriented action $\mathbf{m}^{\mathrm{ref}}_t$ constructed from the reference proxy anchor trajectory.
During training, environments are initialized from reference frames, and the executed action is sampled from a mixed policy that uses the online teacher action with probability $p$ and the student action otherwise.
The student is optimized with PPO using the object-proxy tracking reward and an auxiliary human tracking reward, while an online behavior cloning loss matches the mean action of the teacher.
Both the teacher action replacement probability $p$ and the behavior cloning weight $\lambda_{\mathrm{BC}}$ are gradually annealed to zero, so that the policy shifts from teacher-guided imitation to autonomous proxy-motion execution.
In the second stage, we remove teacher guidance entirely and post-train the same policy with randomly sampled feasible proxy-motion targets $\mathbf{m}^{\mathrm{rand}}_t$.
Only the object-proxy tracking reward is retained in this stage.
This exposes the skill to motions beyond the teacher rollout distribution and improves its robustness before it is frozen for downstream high-level policy learning.

\begin{algorithm}[!t]
\caption{Training the Object-oriented Motion Skill}
\label{alg:low_level_training}
\begin{algorithmic}[1]
\Require Teacher policy $\pi_{\mathrm{tea}}$, reference rollout set $\mathcal{D}_{\mathrm{ref}}$, horizon $K$
\Require Initial low-level policy $\pi_{\mathrm{low}}$, distillation epochs $E_{\mathrm{distill}}$, post-training epochs $E_{\mathrm{post}}$
\Ensure Frozen Object-oriented Motion Skill $\pi_{\mathrm{low}}$

\Statex
\State \textbf{Relabel teacher rollouts as object-oriented supervision}
\For{each rollout $\tau_{\mathrm{ref}}\in\mathcal{D}_{\mathrm{ref}}$}
    \For{each timestep $t$}
        \State Construct
        $\mathbf{m}^{\mathrm{ref}}_t
        =
        [\Delta\mathbf{p}^{\mathrm{ref}}_{t,0},\ldots,
        \Delta\mathbf{p}^{\mathrm{ref}}_{t,K-1}]$
        from the reference proxy anchor trajectory
        \State where
        $\Delta\mathbf{p}^{\mathrm{ref}}_{t,k}
        =
        \mathbf{p}^{\mathrm{ref}}_{t+k+1}
        -
        \mathbf{p}^{\mathrm{ref}}_{t+k}$
    \EndFor
\EndFor

\Statex
\State \textbf{Stage 1: Teacher-guided distillation}
\For{$e=1$ to $E_{\mathrm{distill}}$}
    \State Anneal teacher action probability $p$ and BC weight $\lambda_{\mathrm{BC}}$
    \State Initialize environments from sampled reference frames in $\mathcal{D}_{\mathrm{ref}}$
    \For{each rollout step $t$}
        \State Query student action
        $\mathbf{a}^{\mathrm{low}}_t
        \sim
        \pi_{\mathrm{low}}(\cdot\mid \mathbf{s}_t,\mathbf{o}_t,\mathbf{m}^{\mathrm{ref}}_t)$
        \State Query online teacher action
        $\mathbf{a}^{\mathrm{tea}}_t
        \sim
        \pi_{\mathrm{tea}}(\cdot\mid \mathbf{s}_t,\mathbf{o}_t,\mathbf{g}^{\mathrm{tea}}_t)$
        \State Execute
        $\mathbf{a}_t=\mathbf{a}^{\mathrm{tea}}_t$ with probability $p$,
        otherwise execute $\mathbf{a}_t=\mathbf{a}^{\mathrm{low}}_t$
        \State Compute
        $r_t =
        r^{\mathrm{obj}}_t(\mathbf{m}^{\mathrm{ref}}_t)
        +
        \lambda_{\mathrm{human}} r^{\mathrm{human}}_t$
    \EndFor
    \State Update $\pi_{\mathrm{low}}$ with PPO and online BC:
    \[
    \mathcal{L}_{\mathrm{BC}}
    =
    \mathbb{E}_t
    \left[
    \left\|
    \mu_{\mathrm{low}}(\mathbf{s}_t,\mathbf{o}_t,\mathbf{m}^{\mathrm{ref}}_t)
    -
    \mu_{\mathrm{tea}}(\mathbf{s}_t,\mathbf{o}_t,\mathbf{g}^{\mathrm{tea}}_t)
    \right\|_2^2
    \right].
    \]
\EndFor

\Statex
\State \textbf{Stage 2: Random proxy-motion post-training}
\State Set $p=0$ and $\lambda_{\mathrm{BC}}=0$
\For{$e=1$ to $E_{\mathrm{post}}$}
    \State Sample feasible random proxy-motion targets
    $\mathbf{m}^{\mathrm{rand}}_t$
    \For{each rollout step $t$}
        \State Execute
        $\mathbf{a}_t
        \sim
        \pi_{\mathrm{low}}(\cdot\mid \mathbf{s}_t,\mathbf{o}_t,\mathbf{m}^{\mathrm{rand}}_t)$
        \State Compute
        $r_t =
        r^{\mathrm{obj}}_t(\mathbf{m}^{\mathrm{rand}}_t)$
    \EndFor
    \State Update $\pi_{\mathrm{low}}$ with PPO
\EndFor

\State Freeze $\pi_{\mathrm{low}}$ for downstream high-level policy learning
\end{algorithmic}
\end{algorithm}

During distillation, the student is optimized with the object-proxy tracking reward in Eq.~\eqref{eq:object_tracking_reward}, together with an auxiliary human motion tracking reward:
\begin{equation}
    r^{\mathrm{distill}}_t
    =
    r^{\mathrm{obj}}_t
    +
    \lambda_{\mathrm{human}} r^{\mathrm{human}}_t .
\end{equation}
The object reward enforces execution of the reference object-oriented action $\mathbf{m}^{\mathrm{ref}}_t$, while the auxiliary human reward stabilizes early contact-rich learning by encouraging the student to remain close to feasible teacher motions.
During random post-training, reference human motions and teacher actions are no longer available.
We therefore remove the human tracking term and optimize the low-level skill only with the object-proxy tracking reward under randomly generated proxy-motion conditions:
\begin{equation}
    r^{\mathrm{post}}_t
    =
    r^{\mathrm{obj}}_t,
    \qquad
    \mathbf{m}^{\star}_t=\mathbf{m}^{\mathrm{rand}}_t .
\end{equation}
This prevents the low-level policy from becoming a full-body motion tracker and instead encourages it to realize diverse proxy motions through physically plausible humanoid-object interaction.

Let $\hat{\cdot}$ denote the reference teacher state.
For the auxiliary human reward, we track a compact set of humanoid quantities, including key-point positions $\mathbf{x}^{\mathcal{K}}_t$, body pose $\mathbf{q}_t$, joint velocity $\dot{\mathbf{q}}_t$, and a velocity smoothness term.
The key-point set $\mathcal{K}$ contains the root and arm-related bodies, which are most relevant to stable object interaction.
The pose vector $\mathbf{q}_t$ includes the root orientation and joint poses in the same rotation representation as the policy state.
The human motion tracking reward is defined as
\begin{equation}
\begin{aligned}
    r_t^{\mathrm{human}}
    =
    \exp \Big(
    &-w_{\mathrm{kp}}
    \left\|
    \mathbf{x}^{\mathcal{K}}_t
    -
    \hat{\mathbf{x}}^{\mathcal{K}}_t
    \right\|^2
    -w_{\mathrm{pose}}
    \left\|
    \mathbf{q}_t
    -
    \hat{\mathbf{q}}_t
    \right\|^2 \\
    &-w_{\mathrm{vel}}
    \left\|
    \dot{\mathbf{q}}_t
    -
    \dot{\hat{\mathbf{q}}}_t
    \right\|^2
    -w_{\mathrm{smooth}}
    \left\|
    \dot{\mathbf{q}}_t
    -
    \dot{\mathbf{q}}_{t-1}
    \right\|^2
    \Big).
\end{aligned}
\label{eq:human_tracking_reward}
\end{equation}
This reward is used only as an auxiliary training signal during distillation.
The reference human state is never exposed as part of the low-level policy input.

\paragraph{Procedural Random Proxy-Motion Generation.}
In the post-training stage, the teacher is removed and the goal motion is replaced with a procedurally generated random proxy motion,
$\mathbf{m}^{\star}_t=\mathbf{m}^{\mathrm{rand}}_t$.
The generator maintains a rolling displacement buffer and extends the tail of the current short-horizon trajectory with smooth bounded planar and vertical motions.
Algorithm~\ref{alg:random_proxy_motion} summarizes this procedure.
The generated proxy trajectory is then transformed into the humanoid-centric input frame before being provided to $\pi_{\mathrm{low}}$.

\begin{algorithm}[!t]
\caption{Procedural Random Proxy-Motion Generation}
\label{alg:random_proxy_motion}
\begin{algorithmic}[1]
\Require Current proxy anchor $\mathbf{p}_t$, rolling displacement buffer $\Delta_{t-1}\in\mathbb{R}^{K\times3}$, generator state $\xi_{t-1}$, warm-up length $T_w$
\Ensure Random proxy-motion condition $\mathbf{m}^{\mathrm{rand}}_t$ and updated buffer $\Delta_t$

\If{$t \leq T_w$}
    \State $\mathbf{m}^{\mathrm{rand}}_t \gets \mathbf{0}$, $\Delta_t \gets \mathbf{0}$
    \State Reset generator state $\xi_t$
    \State \Return $\mathbf{m}^{\mathrm{rand}}_t,\Delta_t$
\EndIf

\State Reconstruct the absolute future anchors:
\[
    \mathbf{P}_{1:K}
    =
    \mathbf{p}_t + \mathrm{cumsum}(\Delta_{t-1}) .
\]
\State Let $\mathbf{q}\gets \mathbf{P}_{K}$ be the current trajectory tail.

\If{$t=T_w+1$}
    \State Initialize $\xi_t=(\phi_t,v_t,\zeta_t)$ by sampling yaw, speed, and height-control variables
\Else
    \State $\xi_t\gets\xi_{t-1}$
\EndIf

\State Sample bounded planar increments $\delta\phi$ and $\delta v$
\State Update heading and speed:
\[
    \phi_t \gets \phi_t+\delta\phi,
    \qquad
    v_t \gets \mathrm{clip}(v_t+\delta v).
\]

\State Generate the next planar tail:
\[
    \mathbf{P}^{xy}_{K+1}
    =
    \mathbf{q}^{xy}
    +
    \Delta t\, v_t
    \begin{bmatrix}
    \cos\phi_t\\
    \sin\phi_t
    \end{bmatrix}.
\]

\State Generate the next height:
\[
    P^z_{K+1}
    =
    \begin{cases}
    q^z+\Delta z_{\mathrm{raise}}, & \text{initial lifting stage},\\
    q^z+\Delta z_{\mathrm{smooth}}, & \text{otherwise},
    \end{cases}
\]
where $\Delta z_{\mathrm{smooth}}$ is produced by a low-frequency smooth height process.

\State Clamp $P^z_{K+1}$ to the feasible height range and form
\[
    \mathbf{P}_{K+1}
    =
    [\mathbf{P}^{xy}_{K+1}, P^z_{K+1}].
\]

\State Set the current random proxy-motion condition:
\[
    \mathbf{m}^{\mathrm{rand}}_t
    \gets
    \left[
    \mathbf{P}_{1}-\mathbf{p}_t,\,
    \mathbf{P}_{2}-\mathbf{P}_{1},\,
    \ldots,\,
    \mathbf{P}_{K}-\mathbf{P}_{K-1}
    \right].
\]

\State Update the rolling displacement buffer:
\[
    \Delta_t
    \gets
    \left[
    \mathbf{P}_{2}-\mathbf{P}_{1},\,
    \ldots,\,
    \mathbf{P}_{K+1}-\mathbf{P}_{K}
    \right].
\]

\State \Return $\mathbf{m}^{\mathrm{rand}}_t,\Delta_t$
\end{algorithmic}
\end{algorithm}

\paragraph{Downstream High-Level Policy and Rewards.}
For high-level furniture manipulation, we formulate downstream tasks as object-centric proxy-motion planning over task-relevant manipulation regions.
The initial poses of the humanoids and object, as well as the target object placement, are randomly sampled within predefined task ranges.
The task goal $\mathbf{g}_t\in\mathbb{R}^{3}$ encodes the desired object-level outcome, i.e., the target position of the furniture root in the object-centric frame.
The global object state $\mathbf{o}^{\mathrm{g}}_t$ follows the same $54$D cuboid format as the low-level proxy observation, but is represented in the object-centric coordinate frame.
For a task with $N$ humanoids, each humanoid is assigned a manipulation region $\boldsymbol{\rho}^{i}$ according to the object geometry and its interaction role.
For compact objects, the local proxy can be the whole object; for long or large furniture, different humanoids are assigned to endpoints, sides, or corners; for pushing tasks, the regions are placed on the side from which agents should apply force.
Given $\mathbf{o}^{\mathrm{g}}_t$ and $\boldsymbol{\rho}^{i}$, we instantiate a local cuboid proxy
\begin{equation}
    \mathbf{o}^{i}_t
    =
    \psi(\mathbf{o}^{\mathrm{g}}_t,\boldsymbol{\rho}^{i}),
\end{equation}
which keeps the same observation format as in low-level skill learning.

The high-level policy observes the global object state, the task goal, and the per-agent object-centric inputs
$\{ \bar{\mathbf{s}}^{i}_t,\mathbf{o}^{i}_t \}_{i=1}^{N}$, where each per-agent input has $277$ dimensions.
It outputs one proxy-motion condition $\mathbf{m}^{i}_t\in\mathbb{R}^{30}$ for each humanoid:
\begin{equation}
    \{\mathbf{m}^{i}_t\}_{i=1}^{N}
    =
    \pi_{\mathrm{high}}
    \left(
    \mathbf{o}^{\mathrm{g}}_t,
    \{\bar{\mathbf{s}}^{i}_t,\mathbf{o}^{i}_t\}_{i=1}^{N},
    \mathbf{g}_t
    \right).
\end{equation}
Each high-level output is transformed from the object-centric frame to the corresponding humanoid-centric frame before being executed by the frozen Object-oriented Motion Skill:
\begin{equation}
    \mathbf{a}^{i}_t
    =
    \pi_{\mathrm{low}}
    \left(
    \mathbf{s}^{i}_t,\mathbf{o}^{i}_t,\mathbf{m}^{i}_t
    \right).
\end{equation}

\begin{table}[!t]
\centering
\small
\caption{Implementation details and main hyperparameters.}
\label{tab:implementation_details}
\begin{tabular}{ll}
\toprule
\textbf{Item} & \textbf{Value} \\
\midrule
Simulator & Isaac Gym \\
Number of environments & $4096$ \\
Humanoid model & AMP humanoid \\
Control mode & PD target control \\
Control frequency & $30$Hz \\
Physics substeps & $2$ \\
Episode length & $2500$ steps \\
State initialization & Hybrid initialization \\
\midrule
Policy architecture & MLP actor-critic \\
Hidden units & $[1024,1024,512]$ \\
Activation & ReLU \\
Input normalization & Enabled \\
Value normalization & Disabled \\
Action distribution & Fixed diagonal Gaussian \\
\midrule
Optimizer & PPO \\
Learning rate & $2\times 10^{-5}$ \\
Discount factor $\gamma$ & $0.99$ \\
GAE parameter $\lambda$ & $0.95$ \\
PPO clipping threshold & $0.2$ \\
Rollout horizon & $32$ \\
Minibatch size & $16384$ \\
PPO epochs per update & $6$ \\
Critic loss coefficient & $5.0$ \\
Entropy coefficient & $0.0$ \\
Gradient norm & $1.0$ \\
\midrule
Low-level distillation epochs & $15000$ \\
Low-level random post-training epochs & $5000$ \\
High-level epochs, $N=1$ & $10000$ \\
High-level epochs, $N=2$ & $20000$ \\
High-level epochs, $N=4$ & $40000$ \\
\midrule
Maximum proxy speed & $2.5\,\mathrm{m/s}$ \\
Proxy-motion increment clip & $\pm 0.083\,\mathrm{m}$ per control step \\
\bottomrule
\end{tabular}
\end{table}

During downstream training, the learned Object-oriented Motion Skill is frozen and only the high-level policy $\pi_{\mathrm{high}}$ is optimized.
Following prior physics-based interaction-control methods~\citep{hassan2023synthesizing, gao2024coohoi}, we train $\pi_{\mathrm{high}}$ with PPO using task-specific rewards.
The single-human teacher policy used for low-level distillation is trained independently with the same PPO-based interaction-control protocol.
Its reward follows the same object-transport objective described below, specialized to $N=1$ with the whole box treated as the assigned proxy region.
This makes the teacher rollouts physically feasible single-agent carrying demonstrations that are aligned with the downstream proxy-motion execution objective.

For downstream shared-object manipulation, the high-level task reward consists of a holding term and an object-transport term:
\begin{equation}
    r^{\mathrm{high}}_t
    =
    0.4\, r^{\mathrm{held}}_t
    +
    0.6\, r^{\mathrm{trans}}_t .
    \label{eq:high_level_reward}
\end{equation}
The holding term encourages each humanoid to stay attached to its assigned local proxy region.
Let $\mathbf{h}^{i}_t$ denote the midpoint of the two hands of agent $i$, and let $\mathbf{c}^{i}_t$ denote the geometric center of its assigned local proxy $\mathbf{o}^{i}_t$.
We define the multi-agent holding reward as
\begin{equation}
    r^{\mathrm{held}}_t
    =
    \prod_{i=1}^{N}
    \exp
    \left(
    -w_{\mathrm{held}}
    \left\|
    \mathbf{h}^{i}_t-\mathbf{c}^{i}_t
    \right\|^2
    \right).
    \label{eq:holding_reward}
\end{equation}
The product form requires all agents to maintain their assigned contacts: if any agent loses its proxy region, the overall holding reward decreases sharply.

The object-transport term measures whether the shared object moves toward the target with a proper direction, velocity, distance, and final placement.
Let $\mathbf{p}^{o}_t$ and $\mathbf{v}^{o}_t$ denote the object root position and velocity, and let $\mathbf{p}^{\mathrm{tar}}$ denote the target position.
We compute transport-related terms in the horizontal plane.
Let $\Pi_{xy}(\cdot)$ denote the projection onto the ground plane, and define
\begin{equation}
    \mathbf{d}_t
    =
    \frac{
    \Pi_{xy}(\mathbf{p}^{\mathrm{tar}})-\Pi_{xy}(\mathbf{p}^{o}_t)
    }{
    \left\|
    \Pi_{xy}(\mathbf{p}^{\mathrm{tar}})-\Pi_{xy}(\mathbf{p}^{o}_t)
    \right\|+\epsilon
    },
\end{equation}
where $\mathbf{d}_t$ is the horizontal direction from the object to the target.
The transport reward is
\begin{equation}
    r^{\mathrm{trans}}_t
    =
    0.2\,r^{\mathrm{dir}}_t
    +
    0.2\,r^{\mathrm{vel}}_t
    +
    0.5\,r^{\mathrm{dist}}_t
    +
    0.1\,r^{\mathrm{put}}_t .
    \label{eq:transport_reward}
\end{equation}
The four components are defined as
\begin{align}
    r^{\mathrm{dir}}_t
    &=
    \bar{\mathbf{f}}^{o}_t \cdot \mathbf{d}_t ,
    \label{eq:direction_reward}
    \\
    r^{\mathrm{vel}}_t
    &=
    \exp
    \left(
    -w_{\mathrm{obj,vel}}
    \left(
    \Pi_{xy}(\mathbf{v}^{o}_t) \cdot \mathbf{d}_t - v^{\star}
    \right)^2
    \right),
    \label{eq:velocity_reward}
    \\
    r^{\mathrm{dist}}_t
    &=
    \exp
    \left(
    -w_{\mathrm{dist}}
    \left\|
    \Pi_{xy}(\mathbf{p}^{o}_t)-\Pi_{xy}(\mathbf{p}^{\mathrm{tar}})
    \right\|^2
    \right),
    \label{eq:distance_reward}
    \\
    r^{\mathrm{put}}_t
    &=
    \mathbb{I}
    \left[
    \left\|
    \Pi_{xy}(\mathbf{p}^{o}_t)-\Pi_{xy}(\mathbf{p}^{\mathrm{tar}})
    \right\| < 0.5
    \right]
    \exp
    \left(
    -w_{\mathrm{put}}
    \left(
    z^{o}_t-z^{\mathrm{place}}
    \right)^2
    \right).
    \label{eq:putdown_reward}
\end{align}
Here, $\bar{\mathbf{f}}^{o}_t$ is the normalized horizontal forward direction of the object root, $v^{\star}=1.0\,\mathrm{m/s}$ is the desired transport speed, $z^{o}_t$ is the object root height, and $z^{\mathrm{place}}$ is the expected object root height after final placement.
The direction term is a signed cosine alignment term without an exponential transformation, while the velocity and distance terms provide dense guidance for moving the object toward the target.
The put-down term is activated only when the object is sufficiently close to the target, encouraging correct final placement rather than merely transporting the object near the goal.

\paragraph{Optimization and Hyperparameters.}
All policies are trained in Isaac Gym with $4096$ parallel environments.
We use the same simulated humanoid model and PD-control interface for both low-level and high-level policies.
The control frequency is $30$Hz, with two physics substeps per control step.
Both $\pi_{\mathrm{low}}$ and $\pi_{\mathrm{high}}$ are parameterized by MLP actor-critic networks with hidden sizes $[1024,1024,512]$ and ReLU activations.
The policies output diagonal Gaussian actions with fixed standard deviation, and we optimize them using PPO with generalized advantage estimation.
Table~\ref{tab:implementation_details} summarizes the main implementation details and hyperparameters.

The low-level Object-oriented Motion Skill is trained in two stages.
In the distillation stage, the student is trained for $15000$ epochs using teacher rollouts and the auxiliary human tracking reward.
It is then post-trained for another $5000$ epochs using randomly generated proxy-motion targets, where only the object-proxy tracking objective is used.
For downstream tasks, $\pi_{\mathrm{low}}$ is frozen and only the high-level policy $\pi_{\mathrm{high}}$ is optimized.
We train the high-level policy for $10000$, $20000$, and $40000$ epochs for single-agent, two-agent, and four-agent tasks, respectively.

To stabilize high-level training, we clip the proxy-motion increments predicted by $\pi_{\mathrm{high}}$.
Since the control frequency is $30$Hz, a maximum proxy speed of $2.5\,\mathrm{m/s}$ corresponds to a per-step displacement limit
\begin{equation}
    \Delta p_{\max}
    =
    \frac{2.5}{30}
    \approx
    0.083\,\mathrm{m}.
\end{equation}
We therefore clip each consecutive-frame translational proxy-motion increment to this range.
This prevents the high-level policy from producing physically infeasible proxy trajectories that cannot be executed by the frozen low-level skill.

\begin{figure}[t]
    \centering
    \includegraphics[width=0.86\linewidth]{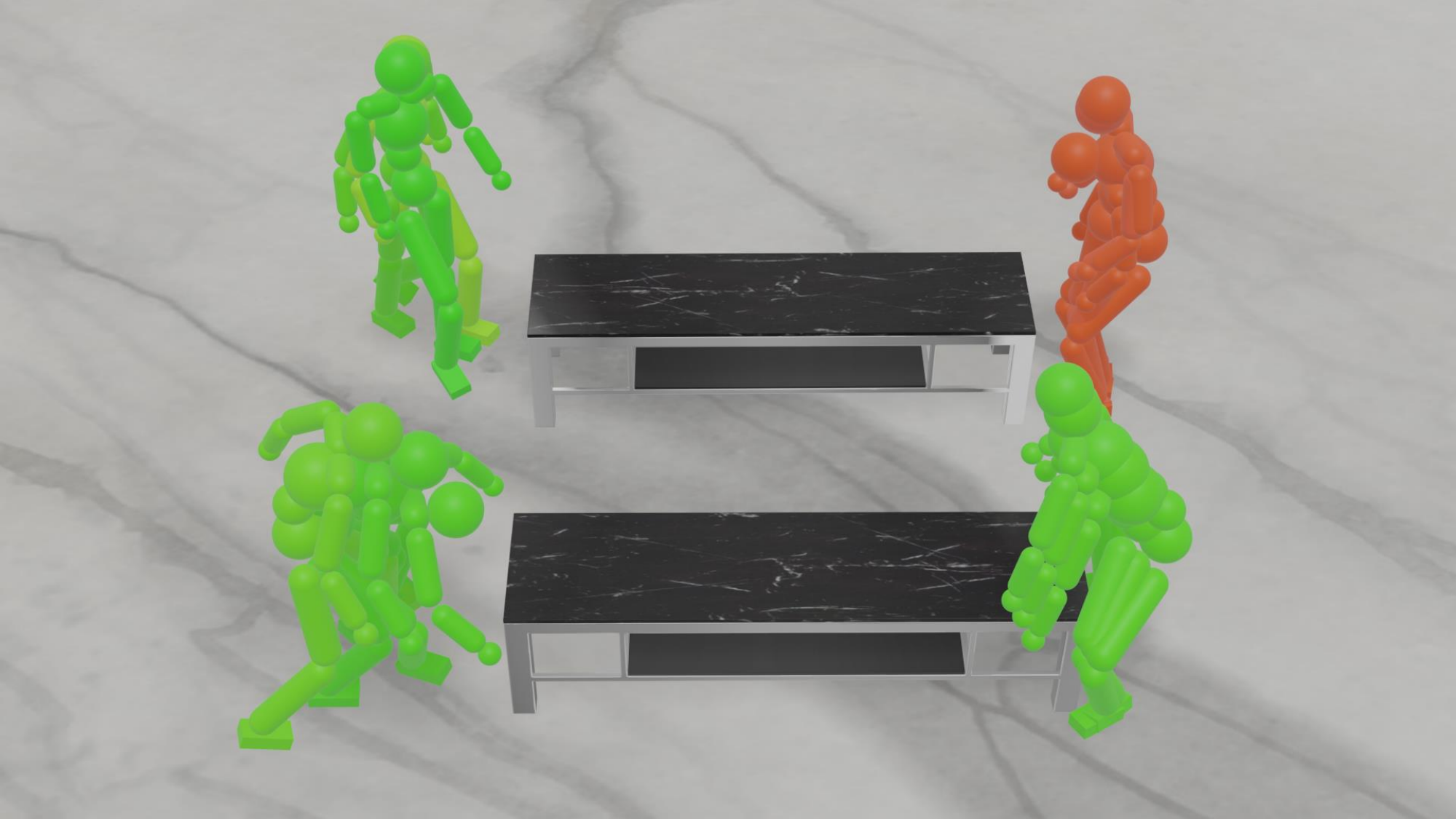}
    \caption{
    A deadlock failure of CooHOI~\citep{gao2024coohoi}. 
    After failed grasping attempts, the agents may stay near the object but stop making progress. 
    Our method is less likely to get stuck because it follows explicit short-horizon trajectory plans.
    }
    \label{fig:compare_coohoi}
\end{figure}

\paragraph{Failure mode compared with CooHOI.}
Fig.~\ref{fig:compare_coohoi} shows a typical failure case of CooHOI~\citep{gao2024coohoi}. 
After one or two failed attempts to lift the object, the policy can get stuck: the agents stay around the object but stop making further effective adjustments. 
In other words, once the interaction falls into a bad local state, CooHOI may fail to recover and the whole cooperative behavior enters a deadlock. 
A likely reason is that CooHOI relies on implicit coordination through object-conditioned observations, which can be sensitive to the local object state, especially object velocity. 
When the agents are already close to the object while the object remains static, the policy may no longer receive a strong enough signal to continue corrective behavior. 
Our method is less prone to this issue. 
The high-level policy explicitly predicts short-horizon local future trajectories, and the low-level skill is further post-trained with RL to respond robustly to diverse trajectory commands. 
As a result, even from imperfect contact states or near-static configurations, our policy can still keep producing meaningful corrective actions instead of becoming inactive.

